\pdfoutput=1

\documentclass[11pt]{article}

\usepackage{emnlp2021}

\usepackage{times}
\usepackage{latexsym}
\usepackage{enumitem,kantlipsum}
\usepackage{graphicx}
\usepackage{caption}
\usepackage{amsmath}
\usepackage{amssymb}
\usepackage{pgfplots}
\pgfplotsset{compat=newest}
\usepackage{comment}
\usepackage{booktabs}
\usepackage{hyperref}
\usepackage[overload]{empheq}
\usepackage{makecell}
\usepackage[ruled,vlined]{algorithm2e}
\usepackage{comment}
\usepackage{pifont}
\usepackage{url}

\newcommand{\specialcell}[2][l]{\begin{tabular}[#1]{@{}l@{}}#2\end{tabular}}

\definecolor{Green}{RGB}{30,148,55}
\newcommand{\Green}[1]{{\color{Green}{#1}}}
\newcommand{\red}[1]{{\color{red}{#1}}}

\newcommand{\cmark}{\ding{51}}%
\newcommand{\xmark}{\ding{55}}%

\usepackage[T1]{fontenc}

\usepackage[utf8]{inputenc}

\usepackage{microtype}

\title{Exploration of the Usage of Color Terms by Color-blind Participants in Online Discussion Platforms}

\author{
Ella Rabinovich\thanks{\hspace{0.07in}The authors contributed equally to this work.} {\hspace{0.05in}}$^{1}$  \hspace{5cm} Boaz Carmeli$^{*} {\hspace{0.05in}}^{1,2}$\\
    ${}^1$ IBM Research \\
    ${}^2$ Technion -- Israel Institute of Technology \\
    \texttt{ella.rabinovich1@ibm.com} \hspace{2cm}
    \texttt{boazc@il.ibm.com} \\
}

\begin{document}
\maketitle

\begin{abstract}
Prominent questions about the role of sensory vs. linguistic input in the way we acquire and use language have been extensively studied in the psycholinguistic literature. However, the relative effect of various factors in  a person's overall experience on their linguistic system remains unclear. We study this question by making a step forward towards a better understanding of the conceptual perception of colors by color-blind individuals, as reflected in their spontaneous linguistic productions. Using a novel and carefully curated dataset, we show that red-green color-blind speakers use the "red" and "green" color terms in less predictable contexts, and in linguistic environments evoking mental image to a lower extent, when compared to their normal-sighted counterparts. These findings shed some new and interesting light on the role of sensory experience on our linguistic system.

\end{abstract}

\section{Introduction}
Colors play an exceptionally prominent role in our lives. Simple and vivid, and yet so difficult to describe or reduce to linguistic terms, our experience of color has long raised intriguing questions concerning  grounded color semantics -- quantifying the associations between words and perceptual representations -- to philosophers, scientists, and psycholinguists \citep{chuang2008probabilistic, heer2012color}. \citet{mcmahan2015bayesian} have shown that the subjective quality of color experience varies between individuals. A body of work in color semantics have indicated that color lexicalization and usage patterns can be significantly affected by extra-linguistic factors, such as culture, physical environment \citep{athanasopoulos2011color, josserand2021environment}, and the native language of a speaker \citep{sarantakis2014influence, matusevych2018crosslinguistic}.

How do colors appear to color-blind individuals? Does the imperfect perceptual experience of red and green in people with red-green visual deficiency (\textit{deuteranopia}) shape their color-related linguistic production? Embodied cognition theory poses that the entirety of our sensory experience -- activities that help us develop a better understanding of word semantics by using our five senses -- shapes our conceptual knowledge \citep{barsalou2008grounded, foglia2013embodied}. As an example, reading the word "cat" is likely to elicit sensory experiences we have with cats, such as their sound and how they look. Embodied cognition theory thus assumes that all our sensory experiences contribute to our conceptual knowledge and processing, which, in turn, is reflected in our language. 

\paragraph{Related Work} Prior work on the effect of color-blindness on language production is relatively sparse. \citet{landau2009language} studied the language of \textit{blind} children, focusing (among others) on the achievements of three blind children's in the area of syntax and word learning. The authors found general development patterns similar to those by their sighted agemates. Representation of colors in blind and color-blind individuals was studied in a controlled color-similarity experiment with 37 participants -- 15 red-green color-blind, among others \citep{shepard1992representation}. The participants were asked to rank the degree of similarity between colors, when presented with names-only, visual colors only, and names+color stimuli. While significant differences in the similarity judgments were found for the color-only setting, when color-deficient participants were presented with names along with the colors, their rankings became closer to those by normal-sighted people. This suggests that linguistic exposure plays a considerable role in shaping our perception of color representation. Multiple works have studied the language of visually impaired and blind children at various stages of language development, suggesting evidence for difficulties in just those areas of language acquisition where visual information can provide input about the world, stimulating hypotheses about pertinent aspects of the linguistic system \citep{andersen1984blind, perez2013language}.

The puzzling question on the role of sensory vs. linguistics input in shaping our color perception remains therefore sound. In this work, we make a step forward towards better understanding of the conceptual perception of the red and green colors in red-green color-blind individuals, as mirrored in their spontaneous linguistic production. 

We perform a first (to the best of our knowledge) large-scale computational study on the usage of the "red" and "green" color terms in (self-reported) population with \textit{deutan} and \textit{protan} visual impairment. Using a novel dataset of linguistic productions by color-blind (CB) individuals, we show that they use the "red" and "green" color terms in less predictable contexts, and in linguistic environments evoking mental image to a lower extent, when compared to normal-sighted (NS) authors.

The contribution of this study is, therefore, twofold: First, we release a large, diverse, and carefully curated dataset of linguistic productions by red-green CB authors, accompanied by a corpus of utterances by NS individuals, aligned on various linguistic properties. Second, we show preliminary evidence for subtle, yet reliably detected, divergences in the usage of "red" and "green" by CB speakers, compared to their NS counterparts. We make the dataset and our code available for facilitating future research in this field.\footnote{Code is available at \url{https://github.com/IBM/colorblind-language}; complying with Reddit's terms of use, we provide a full pipeline for re-producing the dataset (extraction and filtering), rather than the data itself.}

\section{Datasets}
\label{sec:datasets}
We collected datasets used in this work from \href{https://www.reddit.com/}{Reddit} -- an online community-driven platform consisting of numerous forums for news aggregation, content rating, and discussions. As of 2021, it had over 430 million monthly active users, positioning it as the sixth most popular social site in the US. 
Content entries are organized by areas of interest called subreddits, ranging from main forums that receive extensive attention to smaller ones that foster discussion on niche areas. 

\subsection{Collection of Posts by CB Users} 
Multiple subreddits allow their contributors to specify a \textit{flair} -- a metadata attribute adding context to the specific subreddit, such as country of origin, political association, occupation, age, etc. We collected the set of color-blind Reddit authors from \texttt{r/colorblind}, considering only those self-reported as having one of the red-green color blindness types we study in this work: \textit{deuteranopia}, \textit{deuteranomaly}, \textit{protanopia}, and \textit{protanomaly}. This procedure resulted in $2,523$ authors in total. Using the collected list of user IDs, we were further able to retrieve their entire digital footprint from Reddit, spanning years 2005 through 2021. 

Manual inspection of utterances produced by the color-blind Reddit users reveals that CB authors occasionally discuss various aspects related to the impairment, as in \textit{"this game's color-scheme is not a good fit for colorblind, I cannot tell red from green"}. Aiming at the analysis of deficiency-agnostic linguistic productions, we apply strict filters on user utterances, by excluding (1) sentences originating from a manually collected list of subreddits potentially related to the color blindness phenomenon, and (2) sentences containing words possibly indicative of the CB impairment, such as "color", "colorblind", "vision", their inflections and spelling alternatives (e.g., "colour"), to prevent potential biases stemming from deficiency-related discussions. The full list of excluded subreddits can be found in Appendix A.1.

\subsection{Collection of Posts by NS Users}
\label{sec:dataset-nv}
The comparative nature of our analysis requires a collection of utterances produced by normal-sighted Reddit authors. Assuming the relatively low ratio of ${\sim}8\%$ of people with the CB deficiency in the population \citep{wong2011color}, we sampled a large set of posts and comments from the general population of Reddit authors, excluding the (self-reported) set of CB users. We believe that this approach largely targets the language of NS authors due to their large numbers and extensive diversity.

Usage patterns of color terms in linguistic productions can be affected by several dimensions: demographic factors (gender, age), language modality (spoken vs. written), linguistic register (formal vs. informal), topical preferences, etc. Multiple works have shown that there exist detectable differences in the language of male and female speakers, and that topical tendencies shape both the frequency and contextual environment of word usage. Therefore, we strived to create a control set of NS productions that would be aligned with CB language across these dimensions. While achieving a perfect alignment is impractical, we controlled for two major dimensions -- gender and topic -- while sampling linguistic productions by NS authors.

\vspace{-0.01in}
\paragraph{Balancing Posts by Author Gender}
Color blindness affects approximately 1 in 12 men ($8$\%) and 1 in 200 women ($0.5$\%) in the world \citep{wong2011color}. Because most commons roots of color blindness are genetic, passed along the X-chromosome, people with XY chromosomes (most men) only need one defective chromosome (X) to have the deficiency \citep{wong2011color}. Roughly speaking, the phenomenon is $16$ times more frequent in men than in women. The imbalanced $2$:$1$ ratio of male (M) to female (F) Reddit authors\footnote{According to statistics in \url{shorturl.at/doH02}.} imposes an additional prior distribution to the ratio of men vs. women in the color-blind population of Reddit, increasing the estimated frequency of color-blind male authors to be $32$ times higher than that of female in our data.\footnote{The collection of color-blind authors does not contain gender markers; therefore, applying the general Reddit prior to our set of CB authors is a plausible choice.}

A large body of research has shown that the language of female authors differs from that of their male counterparts, exhibiting both topical and stylistic divergences \cite{lakoff1973language, holmes1984women, labov1990intersection}, to the extent that texts written by the two genders are separable automatically \citep{koppel2002automatically, argamon2003gender, rabinovich2017personalized}. Gender-linked differences in human color lexicon, preferences, and perception have been reported in the literature \citep{arthur2007gender, eckert2013language}, suggestive of their effect on both the frequency and contextual linguistic environments of color terms. A valid control set of authors should, therefore, maintain the same M:F author ratio as in the CB set, i.e., $32\text{:}1$.

Recently, \citet{rabinovich2020pick} released a large dataset of posts and comments collected from the Reddit discussion platform, where each sentence is annotated by the (self-reported) binary author gender. We exploit this dataset by making use of utterances by $13,630$ male users, and by (randomly downsampled) $425$ female users, preserving the $32\text{:}1$ M:F author ratio and resulting in the total of $14,055$ authors\footnote{Authors with self-reported gender that also indicated their color blindness defect, were excluded from this set.} and over $45$M posts. 

\vspace{-0.01in}
\paragraph{Balancing Posts by Topical Threads}
Usage patterns of words, and in particular, color terms, are likely to be affected by their contextual environment. As an example, using color terms in a topical thread (subreddit) related to interior design will differ from that of gaming, health, or world news. 

Aiming at similar topical distribution in both CB and NS sets, we balance the distribution of sentences in various subreddits across the two populations, by (1) splitting the data at the sentence-level, (2) using the CB subreddit distribution as the anchor, and (3) performing \textit{stratified sampling} of NS data to maintain the same relative topical ratios. Specifically, let $\mathcal{R}${=}$(r_1, r_2, r_3, ..., r_n)$ be the relative ratios of the amount of sentences spanning $n$ subreddits in the CB dataset, where $\sum{r_i}{=}1$; the set of NS sentences is then randomly downsampled in a manner preserving the topical distribution $\mathcal{R}$. Although the absolute number of sentences differs significantly in the two datasets, the relative ratio of each topical thread is roughly preserved.

\subsection{Color Terms used in this Study}
We address our research questions by performing contrastive analysis of the usage patterns of "red" and "green", as well as additional eight color terms exceeding the total count of $1000$ in our CB dataset: "black", "white", "blue", "brown", "gr[ae]y", "yellow", "pink" and "purple".\footnote{With an exception of "purple" that has 943 occurrences.} This resulted in the total number of over $80$K and $380$K sentences, each including at least one of the ten color terms, for the CB and NS populations, respectively. Differences (if they exist) are anticipated to be linked to the CB-deficiency, therefore evident in the usage of "red" and "green" terms, but not the others.

\subsection{Fixed Expressions and Named Entities}
\label{sec:fixed-exp}
Color terms are often used in fixed linguistic expressions -- groups of words used together to express a particular idea or concept that is more specific than the literal combination of individual words. Among such expressions are "black music", "red army", "green energy", etc. Both the production and comprehension of such expressions is unlikely to evoke a visual image of color in one's mind, hence processing of these terms does not rely on the ability to visually distinguish between colors. Therefore, we excluded expressions with salient non-compositional nature from this work. 

A subset of expressions exceeding the $0.5$\% relative frequency among the full set of <color-term NOUN> adjective phrases considered in this work was examined by a native English speaker. Out of $220$ unique expressions, $140$ were marked as having a common fixed reading, or referring to named entities, such as sport teams ("Green Bay", "Blue Jays"), bands ("Green Day"), or video games ("Red Redemption"). This procedure resulted in excluding about $25$\% of sentences; the complete list of excluded expressions can be found in Appendix A.2.
Table~\ref{tbl:datasets} presents the statistics of our final dataset, spanning over $30$K subreddits. We also report the statistics of two complementary CB datasets released with this work: the collection of posts by authors with blue-yellow color blindness (\textit{tritanopia}) and monochrome vision (\textit{achromatopsia}), to facilitate further research in this field.

\begin{table}[hbt]
\centering
\resizebox{\columnwidth}{!}{
\begin{tabular}{l|rr}
dataset & users  & \# sent (\# with a color term) \\ \hline
protan (CB) & 1,067 & 4.1M (24K) \\
deutan (CB) & 1,456 & 6.0M (36K) \\ \hline
total red-green CB & 2,523 & 10.1M (60K) \\ \hline \hline
normal-sighted (NS) & 14,055 & 45.7M (280K) \\ \hline \hline
tritan (CB) & 236 & 386K (3.8K) \\
monochrome (CB) & 47 & 100K (589) \\
\end{tabular}
}
\caption{Details of the datasets. \# of sentences including one of the color terms is in parentheses. Red-green CB and NS are used in this work; the additional tritan and monochrome datasets are released as well.}
\label{tbl:datasets}
\end{table}

\section{Research Questions}
\label{sec:research-questions}
The two sub-corpora represent a suitable testbed for investigating questions about the unique linguistic phenomena characteristic of red-green CB authors, compared to the NS population of Reddit users. Here, we elaborate on the research questions addressed in this study.

\paragraph{RQ1} 
How does the frequency of color terms in linguistic productions of CB users compare to that of NS speakers? We refer to (1) the frequency of color terms in the language, and (2) the relative frequency ratio of individual color terms -- in particular, "red" and "green" -- within the entire set of the ten color terms considered in this work.

\paragraph{RQ2} 
Red-green color blindness affects the ability to generate a clear (and distinguishable) mental image of these two colors in the mind of a speaker, giving rise to the hypothesis that CB authors would be more hesitant when using these two color terms in linguistic environments evoking a visual image in one's mind. Such linguistic environments can be commonly found in topical threads involving visual experience, such as \texttt{r/gaming}, \texttt{r/nature} or \texttt{r/fashion}. Focusing on \textit{adjective phrase}s with color-terms -- \textit{attributive} (e.g., "red/ADJ shirt/NOUN") or \textit{predicative} (e.g., "this shirt/NOUN is red/ADJ") -- we test this hypothesis by searching for detectable differences in the psycholinguistic property of the modified nouns' \textit{imageability} -- a measure of how easily a physical object, word, or environment evokes a clear mental image in the mind of a person observing it \citep{cortese2004imageability, scott2019glasgow}. 

A common way to study perceptual aspects related to language in psycholinguistic literature distinguishes between nine major psycholinguistic dimensions, including imageability. \citet{scott2019glasgow} released a set of $5,500$ English words manually ranked along the nine dimensions on the $1\text{-}7$ scale, facilitating much research in psycholinguistics and related fields \citep{lewis2020gender, lynott2020lancaster, rabinovich2020pick}. As a concrete example, the word "piano" has a ranking of $6.88$ in the imageability dimension, while "request" was only assigned the score of $2.50$.

We use the rankings by \citet{scott2019glasgow} to investigate if detectable differences can be found in the imageability properties of nouns modified by the "red" and "green" color terms, as employed by red-green CB vs. NS authors. 

\paragraph{RQ3}
Multiple factors influence our lexical choices. Linguistic evidence, extra-linguistic experience, and psychological factors affect the way we employ various linguistic devices in a context. Permanent lack of or deficiency in a sensory input may influence our word usage \citep{andersen1984blind, perez2013language}. One such effect can potentially be manifested by more \textit{conservative} or, on the contrary, more \textit{atypical} usages of a linguistic device. Considering the impaired visual experience in CB users, here we ask if the contextual usage of the "red" and "green" color terms differs between the two populations.

We investigate this question by quantifying the \textit{contextual predictability} of various color terms in the two populations. Contextual predictability of a linguistic unit defines how probable it is in some local environment, thereby providing a way to estimate the differences in the likelihood of color terms in that given context. Higher predictability would be indicative of more common usage patterns; lower predictability -- of less typical choices.

\section{Experiments}
\subsection{Experimental Setup}
We test the suggested research questions on the usage of the "red" and "green" color terms, and compare the findings to usage patterns of the additional eight color terms, as listed in Section~\ref{sec:datasets}. We strengthen these findings by performing similar comparative tests on two control sets, where we do not anticipate differences between CB and NS. All differences were tested for significance, where Bonferroni correction was applied with $m\text{=}20$.

\paragraph{Control Set \#1: Matched Adjectives}
Focusing on the most common syntactic role of color terms in the English language (almost $80$\% of all color terms are tagged as ADJ in the \href{https://nlp.lsi.upc.edu/wikicorpus/}{POS-tagged Wikipedia dump}), we apply the same set of experiments on ten adjectives matched on frequency and length ($\pm1$ character) with the ten color terms. The adjectives include "hot", "flat", "social", "clear", "tiny", "loose", "lame", "petty", "clever", "royal". Differences detectable in "red" and "green", but not in this control set, would be indicative of phenomena unique to "red" and "green" in CB vs. NS authors.

\paragraph{Control Set \#2: NS Authors Random Split}
As an additional control set, we perform a random split of all normal-sighted users preserving the CB:NS user ratio similar to that in the main corpora, as well as gender and stratified topical balance (see Section~\ref{sec:dataset-nv} for details). Differences evident in the "red" and "green" terms in the CB vs. NS main datasets, but not in the NS1:NS2 random split, would imply that they cannot be attributed to random effects. We report detailed experimental results for this control set in Appendix A.3.

\paragraph{Preprocessing}
All posts were split at the sentence level and tokenized using the \href{https://spacy.io/}{spacy} toolkit. Sentences shorter than $4$ or longer than $50$ tokens were excluded from the analysis, as were sentences with a single token longer than $50$ characters.

\subsection{Experimental Results}
\subsubsection{RQ1 -- Frequency and Relative Ratio}
We extracted relative frequencies of all color terms and control adjectives (control set \#1) in the two sub-corpora, along with their frequency ratios. Table~\ref{tbl:frequencies} presents the results. Significant differences between the two populations exist (in both directions), but they are not restricted to the red-green terms. Control set \#2 split yielded differences in a single color term and an adjective (see Appendix A.3). We conclude that no outright CB-linked differences can be found in the frequency-related usage of the two terms in our data.

\begin{table}[h!]
\centering
\resizebox{\columnwidth}{!}{
\begin{tabular}{l|rl|rr}
color term & CB freq  & NS freq & CB ratio  & NS ratio \\ \hline
red     & 7.87e-5 & 7.72e-5     & 0.204 & 0.203 \\
green   & 3.77e-5 & 3.49e-5*    & 0.098 & 0.092 \\ \hline
black   & 8.50e-5 & 8.91e-5*    & 0.222 & 0.235 \\
white   & 6.93e-5 & 7.24e-5     & 0.183 & 0.191 \\
blue    & 4.77e-5 & 4.31e-5*    & 0.124 & 0.114 \\
brown   & 1.92e-5 & 2.02e-5     & 0.051 & 0.053 \\
gr[ae]y & 4.69e-5 & 4.30e-5     & 0.037 & 0.032 \\
yellow  & 1.41e-5 & 1.29e-5     & 0.034 & 0.034 \\
pink    & 9.37e-6 & 8.67e-6     & 0.024 & 0.024 \\
purple  & 8.74e-6 & 8.17e-6     & 0.023 & 0.022 \\ \hline
\textbf{total} &  &  & 1.0 & 1.0 \\ \hline \hline
hot     & 1.07e-4 & 1.11e-4*    & 0.206 & 0.206 \\
social  & 1.12e-4 & 1.14e-4*    & 0.239 & 0.263 \\
clear   & 1.12e-4 & 1.12e-4     & 0.235 & 0.223 \\
tiny    & 5.26e-5 & 5.09e-5     & 0.101 & 0.096 \\
flat    & 4.66e-5 & 4.47e-5     & 0.089 & 0.083 \\
loose   & 2.20e-5 & 2.19e-5     & 0.042 & 0.040 \\
petty   & 1.20e-5 & 1.19e-5     & 0.023 & 0.022 \\
clever  & 1.29e-5 & 1.36e-5     & 0.024 & 0.025 \\
royal   & 1.07e-5 & 1.18e-5     & 0.021 & 0.022 \\
lame    & 1.03e-5 & 1.07e-5     & 0.020 & 0.020 \\ \hline
\textbf{total} &  &  & 1.0 & 1.0
\end{tabular}
}
\vspace{-0.1in}
\caption{Relative frequencies (left) and relative ratios (right) of color terms in the language of CB and NS authors. Statistical significance of the differences was tested using a two-proportion z-test; "*" indicates significant difference at the level of $p\text{<}.01$.}
\label{tbl:frequencies}
\end{table}

\subsubsection{RQ2 -- Imageability of Modified Nouns}
Given a sentence with a color term, we extract the noun modified by the term (where it exists) by applying dependency parsing\footnote{We make use of the spacy POS-tagger and dependency parser for this purpose: \url{https://spacy.io/}} and detecting dependencies connecting the color term as ADJ to a NOUN via the AMOD dependency type, capturing both attributive and predicative adjective phrases. To eliminate spelling mistakes and parsing inaccuracies, we restrict the extracted noun set to the top-$20,000$ most frequent nouns in the corpus.

\begin{table}[hbt]
\centering
\resizebox{\columnwidth}{!}{
\begin{tabular}{lcr|lcr}
noun & orig & score  & noun & orig & score \\ \hline
apple & \cmark & 0.99 & economy & \cmark & 0.07 \\
helicopter & \cmark & 0.98 & philosophy & \cmark & 0.09 \\
cabbage & \xmark & 0.91 & concern & \xmark & 0.19 \\
cobra & \xmark & 0.93 & purity & \xmark & 0.18 \\
\end{tabular}
}
\vspace{-0.1in}
\caption{Example word imageability scores ("orig" denotes scores retrieved from \citet{scott2019glasgow}). 
}
\label{tbl:imageability-ranks}
\end{table}

\begin{table*}[h!]
\centering
\resizebox{\textwidth}{!}{
\begin{tabular}{c|l|c|r}
type & sentence (verbatim)  & noun & img score \\ \hline
A & His \red{\textbf{red}} \textbf{shirt} looks a little derpy & shirt & 0.95 \\
P & I could clearly see that the car didn't stop although the \textbf{light} was \red{\textbf{red}} & light & 0.75 \\ \hline
A & typically, \Green{\textbf{green}} \textbf{bars} and pixellation are a sign that a graphics card is crashing & bars & 0.94 \\
A & Wow... Can't believe we're going to have a \Green{\textbf{green}} \textbf{Christmas} again & Christmas & 0.70 \\
\end{tabular}
}
\vspace{-0.1in}
\caption{Example sentences with color terms in adjective phrases, along with the modified nouns' imageability scores. "P" stands for predicative and "A" for attributive adjective phrase.}
\label{tbl:imageability-examples}
\end{table*}

\begin{figure*}[h!]
\centering
\resizebox{0.87\textwidth}{!}{
\includegraphics{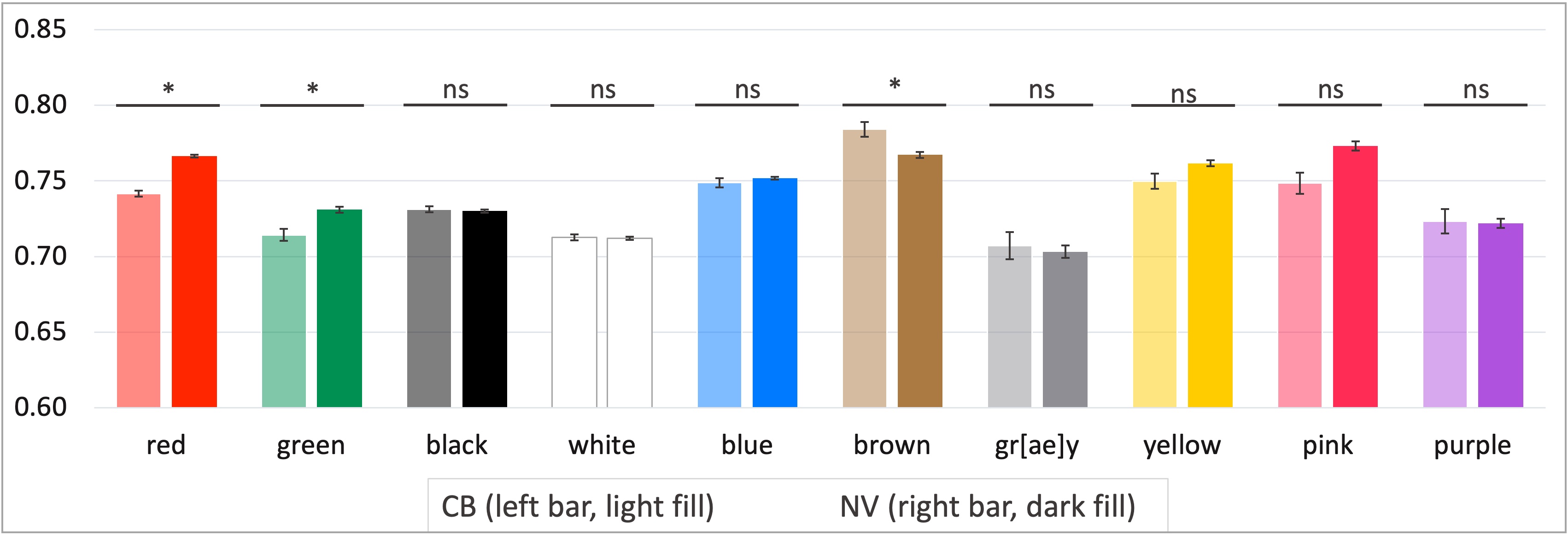}
}
\caption{Mean and standard error of imageability scores of nouns modified by the color terms adjectives. \\ "*" indicates significant difference at the level of $p\text{<}.01$; "ns" indicates non-significant difference. Note the large sample sizes for red and green (contributing to the significance of the findings), but much smaller samples for pink and yellow (Table \ref{tbl:all-img}). Considering the relatively high effect size for pink (Table \ref{tbl:all-img}), the high difference in mean scores would likely be triggered significant for sufficiently large data.}
\label{fig:color-img}
\end{figure*}

\begin{figure*}[h!]
\centering
\resizebox{0.87\textwidth}{!}{
\includegraphics{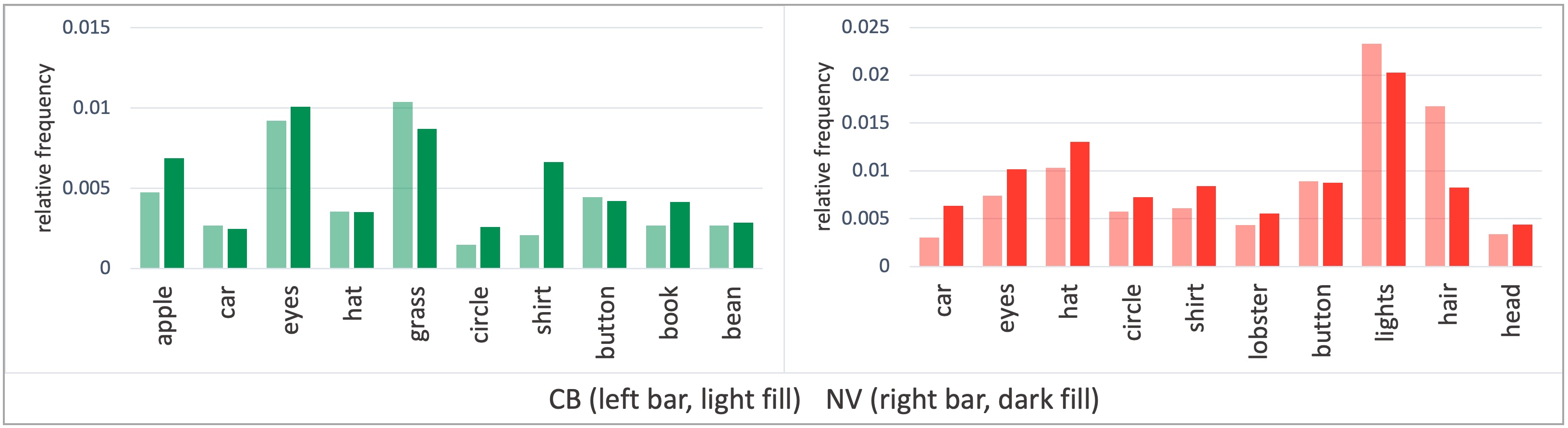}
}
\caption{Top-$10$ most-imageable nouns used with "green" (left) and "red" (right) color terms in attributive and predicative expressions, and their relative frequencies in the CB (left, light fill) and NS (right, dark fill) datasets.}
\label{fig:color-img-freq-nouns}
\end{figure*}

\begin{table}[h!]
\centering
\resizebox{\columnwidth}{!}{
\begin{tabular}{l|rr|rr|r}
& \multicolumn{2}{c|}{CB} & \multicolumn{2}{c|}{NS} & \\ \hline
term & \# sent      & M(img) & \# sent  & M(img) & Cohen's d \\ \hline
red     & 8,249     & 0.744     & 37,213     & 0.768*               & 0.119 \\
green   & 3,267     & 0.716     & 14,823     & 0.733+               & 0.076 \\ \hline
black   & 9,462     & 0.733     & 45,763     & 0.731\hspace{0.20cm} & -0.008 \\
white   & 7,450     & 0.715     & 36,669     & 0.713\hspace{0.20cm} & -0.007 \\
blue    & 4,766     & 0.751     & 20,056     & 0.753\hspace{0.20cm} & 0.010 \\
brown   & 1,292     & 0.788     & 6,227      & 0.769*               & -0.094 \\
gr[ae]y & 1,125     & 0.707     & 4,237      & 0.703\hspace{0.20cm} & -0.051 \\
yellow  & 1,380     & 0.757     & 5,936      & 0.766\hspace{0.20cm} & 0.048 \\
pink    & 846       & 0.75      & 3,915      & 0.775\hspace{0.20cm} & 0.120 \\
purple  & 796       & 0.726     & 3,753      & 0.723\hspace{0.20cm} & -0.014 \\ \hline \hline
hot     & 10,700    & 0.747     & 50,151     & 0.749\hspace{0.20cm} & 0.007 \\
social  & 16,220    & 0.442     & 84,819     & 0.443\hspace{0.20cm} & 0.005 \\
clear   & 6,295     & 0.530     & 27,110     & 0.531\hspace{0.20cm} & 0.004 \\
tiny    & 6,796     & 0.608     & 30,014     & 0.608\hspace{0.20cm} & 0.001 \\
flat    & 3,789     & 0.676     & 16,462     & 0.669\hspace{0.20cm} & -0.029 \\
loose   & 1,480     & 0.646     & 6,846      & 0.641\hspace{0.20cm} & -0.020 \\
petty   & 998       & 0.513     & 4,818      & 0.505\hspace{0.20cm} & -0.045 \\
clever  & 1,130     & 0.492     & 5,424      & 0.483\hspace{0.20cm} & -0.042 \\
royal   & 1,147     & 0.691     & 5,812      & 0.694\hspace{0.20cm} & 0.015 \\
lame    & 930       & 0.524     & 4,355      & 0.536\hspace{0.20cm} & 0.053 \\

\end{tabular}
}
\caption{Mean imageability scores of nouns used with color terms and the control set \#1 adjectives. "*" indicates significant difference at the level of $p\text{<}.01$, "+" indicates significant difference at the level of $p\text{<}.05$.}
\label{tbl:all-img}
\end{table}

\begin{table*}[h!]
\centering
\resizebox{\textwidth}{!}{
\begin{tabular}{l|r}
sentence (verbatim)  & predicted rank \\ \hline
shows all enemies with a \red{\textbf{red}} square on them, it makes it easy to see enemies behind trees, ... & 1 \\
Well if it works then more \red{\textbf{red}}! get those chucks on and be ready :joy: & 809 \\ \hline
Bare feet on \Green{\textbf{green}} grass, especially after a long day of having to wear shoes. & 17 \\
Nothing but the best for our boys in \Green{\textbf{green}}. & 362 \\
\end{tabular}
}
\vspace{-0.1in}
\caption{Example sentences with and BERT rank assigned to the (masked) red-green color term.}
\label{tbl:rank-examples}
\end{table*}

\begin{table}[h!]
\centering
\resizebox{\columnwidth}{!}{
\begin{tabular}{l|rr|rr|r}
& \multicolumn{2}{c|}{CB} & \multicolumn{2}{c|}{NS} & \\ \hline
term & \# sent      & M(rank) & \# sent  & M(rank) & Cohen's d \\ \hline
red     & 11,332    & 109.1     & 50,801    & 94.8*                 & 0.074 \\
green   & 5,734     & 163.6     & 24,208    & 139.3*                & 0.109 \\ \hline
black   & 12,850    & 74.2      & 62,379    & 72.3\hspace{0.20cm}   & 0.013 \\
white   & 10,514    & 99.0      & 51,741    & 95.6\hspace{0.20cm}   & 0.019 \\
blue    & 7,057     & 142.6     & 29,311    & 135.1\hspace{0.20cm}  & 0.034 \\
brown   & 2,900     & 161.5     & 14,028    & 148.2\hspace{0.20cm}  & 0.056 \\
gr[ae]y & 1,692     & 225.1     & 6,969     & 213.2\hspace{0.20cm}  & 0.043 \\
yellow  & 1,900     & 154.8     & 7,885     & 148.3\hspace{0.20cm}  & 0.029 \\
pink    & 1,177     & 186.1     & 5,355     & 149.7*                & 0.156 \\
purple  & 943       & 227.0     & 4,465     & 223.8\hspace{0.20cm}  & 0.012 \\ \hline \hline
hot     & 16,362    & 64.5      & 79,978    & 60.6\hspace{0.20cm}   & 0.027 \\
social  & 16,897    & 32.4      & 88,476    & 27.4*                 & 0.050 \\
clear   & 20,149    & 51.8      & 89,361    & 48.9*                 & 0.023 \\
tiny    & 7,486     & 75.4      & 33,297    & 75.1\hspace{0.20cm}   & 0.002 \\
flat    & 6,909     & 96.3      & 30,898    & 91.0\hspace{0.20cm}   & 0.029 \\
loose   & 3,224     & 141.2     & 14,880    & 137.5\hspace{0.20cm}  & 0.016 \\
petty   & 1,516     & 156.6     & 7,428     & 159.5\hspace{0.20cm}  & -0.013 \\
clever  & 2,300     & 122.9     & 10,830    & 108.2\hspace{0.20cm}  & 0.079 \\
royal   & 1,352     & 100.6     & 7,003     & 89.4\hspace{0.20cm}   & 0.059 \\
lame    & 1,653     & 185.3     & 8,047     & 189.0\hspace{0.20cm}  & -0.016 \\

\end{tabular}
}
\vspace{-0.1in}
\caption{Mean BERT rank predictions for a masked term. Results for both color terms and control set \#1 adjectives are reported. "*" indicates significant difference at the level of $p\text{<}.01$.}
\label{tbl:cwp}
\end{table}

\vspace{-0.075in}
\paragraph{Inferring Missing Imageability Rankings}
The imageability rankings in \citet{scott2019glasgow} cover about $3,000$ of the $20,000$ unique nouns identified in our CB and NS datasets. We exploit these ranking for supervision in extracting ratings for additional nouns in this work. Word embedding spaces have been shown to capture variability in affective dimensions \citep{hollis2016principals} and word concreteness \citep{tsvetkov2013cross, francis2021quantifying}, where imageability is highly correlated with word concreteness \citep{scott2019glasgow}. These findings imply that such semantic representations carry over information useful for the task of assessment of psycholinguistic properties. 

We first normalized the imageability scores in \citet{scott2019glasgow} into the 0\text{-}1 range for better interpretability. Using distributional word representations for the $5,500$ annotated words, we trained a beta regression model\footnote{An alternative to linear regression in situations where the dependent variable is a proportion (0\text{-}1 range).} to predict imageability scores from word embeddings. We further used the trained model to infer imageability rankings for the unlabeled set of nouns. 
We used the \textit{fasttext} word representations \citep{bojanowski2016enriching}, obtaining the highest Pearson's correlations of $0.76$ with the human annotated ratings on a held-out set of $500$ nouns.\footnote{Slightly lower correlations of $0.75$ and $0.68$ were obtained with word2vec \citep{mikolov2013efficient} and Glove \citep{pennington2014glove} embeddings, respectively.}
Table~\ref{tbl:imageability-ranks} presents a sample of nouns with contrasting imageability ratings -- both original and inferred by the regression model.

\paragraph{Assessing the Differences in Imageability Scores of Modified Nouns}
Next we estimated the differences across the imageability dimension in CB vs. NS authors, by recording the imageability score of modified nouns (where it exists) in the productions of the two populations. Table~\ref{tbl:imageability-examples} presents example sentences with "red" and "green" terms and their modified nouns' imageability score in our dataset. We construct two lists of imageability scores: one for the nouns of CB speakers, and another for NS authors. Wilcoxon ranksum significance test was applied to the CB/NS pair of series of values, testing for significant difference, and Cohen's-$d$ was calculated to indicate the magnitude of the effect.

\paragraph{Results and Discussion} Figure~\ref{fig:color-img} presents the differences in the mean imageability score of modified nouns, and Table~\ref{tbl:all-img} reports the full results. Significant differences exist for the "red" and "green" terms, where higher average imageability is observed in the NS authors, suggestive of less frequent use of these color terms to describe entities evoking a clear mental image in a speaker's mind by CB users. The opposite difference is evident for the brown color, possibly indicative of the compensatory usage of "brown" by CB users with high-imageability nouns. The relatively low effect size -- $0.119$, $0.076$ and $-0.094$ for "red", "green", and "brown", respectively -- imply subtle (yet reliably detected) differences in the two populations. 
No control set \#1 adjectives exhibit significant differences, implying that the phenomenon is limited to color-term usage. The same experiment with the control set \#2 yielded no significant differences, with an exception of "black" (with very low Cohen's-$d$\text{=}$0.038$) and "royal" (Cohen's-$d$\text{=}$-0.093$), which can be attributed to subtle topical differences in our data (Appendix A.3).

Figure~\ref{fig:color-img-freq-nouns} presents the top-$10$ most-imageable nouns used with "green" and "red" color terms in attributive and predicative expressions, and their relative frequencies in the CB and NS datasets. Out of ten nouns used in expressions with "green", CB authors use six less frequently than their NS counterparts do, and, similarly, seven out of ten nouns modified by "red". Note the possibly fixed meaning of "green car" (environmentally friendly) -- an expression that is used slightly more frequently by the CB authors. Collectively, these results are suggestive of the less common use of adjective phrases including the "red" and "green" color terms with high-imageability nouns.

\subsubsection{RQ3 -- Contextual Predictability}
Recent advances in deep neural networks (DNN) \citep{lecun2015deep, bengio2021deep}, and specifically, in training large DNN-based masked language models (MLM) \citep{vaswani2017attention} to predict the most plausible word given a sentence context -- a technique known as context-based word prediction \citep{tenney2019you} -- offer novel ways to quantify differences in word usage while comparing large text corpora.
Contextual word prediction (CWP) is a task involving two steps: First, a DNN-based language model is trained over large text corpora to predict masked words within sentence context. Second, the pretrained MLM is used to predict the probability of a word to appear in a specific masked position within a given sentence. 

In this work, we use CWP to compare the "predictable" usage of color terms in language authored by CB and NS populations. For a set of sentences containing color terms (or control set \#1 adjectives), we make use of BERT \citep{devlin2018bert} -- an MLM pretrained on large text corpus to assess the predictability of a term in a given context.\footnote{We make use of the "bert-large-uncased" model; all words in this study were represented by a single token.} We investigate RQ3 by comparing the average CWP scores for color and adjective terms under test in the CB and NS population of Reddit authors. Higher mean predictability would be indicative of a more typical way a term is used in its context.

\paragraph{Assessing Color Terms Predictability}
We predict the probability of a term to appear in a sentence context using BERT \citep{devlin2018bert}. Due to the low ratio of CB people in the general population, we assume that the model broadly mirrors linguistic patterns of normal-sighted people. As a result, better predictability of a word by the model implies a more common usage.

We perform the following steps: (1) mask the designated term (either a color term or an adjective) using the <MASK> token provided by the BERT vocabulary; (2) use the probability distribution over the lexicon produced by the model as a prediction of the masked token. The prediction can be manifested in two ways: (a) the \textit{probability value} produced by the softmax layer, and (b) the \textit{rank} inferred by the probability distribution over the lexicon. For better interpretability, we make use of the latter (rank). Table~\ref{tbl:rank-examples} presents example sentences with red-green color terms and the BERT rank associated with the prediction of the (masked) term: "red" or "green".

Exceptionally high ranks were assigned to sentences with very atypical usage patterns of color terms, in particular, typos and ungrammatical usages. As a concrete example, the color term "red" in the sentence \textit{"that was how i \textbf{red} it at first"} was assigned the rank of $5573$ by the model since it is a typo, making it very unpredictable in this context. We therefore filtered out all sentences with a rank exceeding $1000$ from this computation, as such rare cases significantly affect the average rank prediction.\footnote{The rank of $1000$ was selected by qualitative evaluation over the set of $100$ sentences with color terms.} This filtering approach reduces the total number of color-terms sentences by $9.3\%$ and $8.2\%$, as well as control set \#1 sentences by $5.6\%$ and $4.9\%$, produced by the CV and NS authors, respectively.
The suggested approach results in two lists of ranks for CB and NS productions. We further test the two lists for significant difference, and calculate Cohen's-$d$ effect size.

\begin{figure}[h!]
\centering
\resizebox{\columnwidth}{!}{
\includegraphics{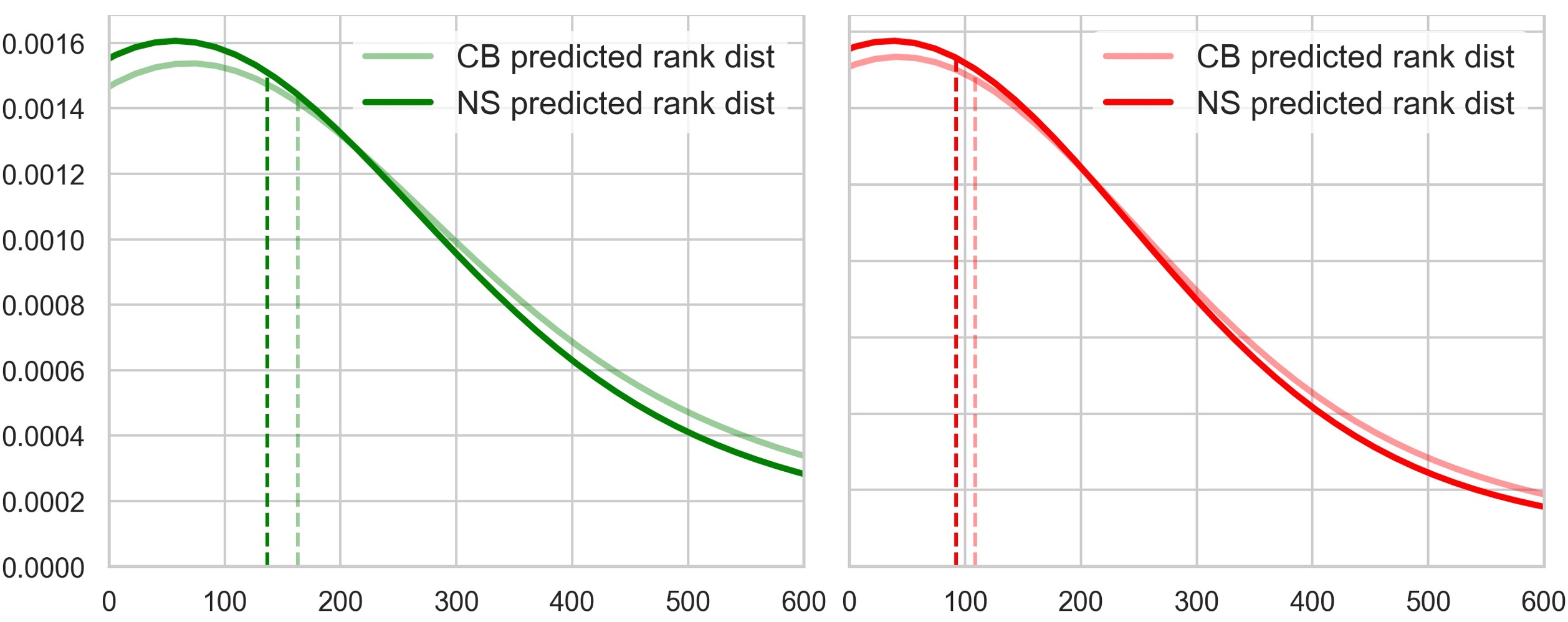}
}
\caption{KDE plot of predicted rank for the red and green color terms in the CB vs. NS population; x-axis is pruned at $600$; dashed line indicates the mean value.}
\label{fig:ranks-kde}
\end{figure}

\paragraph{Results and Discussion} 
Table~\ref{tbl:cwp} reports the results.
Evidently, "green", "red", and "pink" are the three terms exhibiting the highest Cohen's-$d$ scores among all terms with significant differences. Interestingly, none of the adjective and other color terms with significant differences exceed the effect size of $0.05$. The differences in "pink" could possibly be attributed to its proximity to the red color. Among the adjectives, only "social" and "clear" show significant difference, with very low effect size. Figure~\ref{fig:ranks-kde} illustrates the kernel density estimation (KDE) of predicted ranks for the green and red color terms in CB vs. NS speakers: the lower density for most predictable usages (around the rank of $0$) and the slight right shift indicate less typical usage patterns of CB authors. No significant differences were found for colors in the control \#2; detailed results are presented in Appendix A.3.

\section{Discussion}
We view the main contribution of this study in its large-scale data-driven empirical evidence for theoretically-motivated hypotheses on the effect of various sensory experiences on language learning and linguistic production. Human language production is a complex cognitive skill, where most psychological models agree on three stages: conceptualization, formulation and articulation. Conceptualization is considered to be the phase of selection and preparation of pre-linguistic information, relying also on extra-linguistic knowledge – among others, visual perception based on our sensory experience. LLMs, however large and complex, (presumably) lack this inherently-human cognitive skill, but rather operationalize linguistic production by stochastically reproducing language constructions they were exposed to, and generalizing to additional linguistically plausible patterns. Focusing on the effect of multi-modal input on human language production, we ask if bridging this gap is required for contemporary LLMs in order to generate fully naturalistic, human-like language. 
We believe that this work sheds new and interesting light on one of the core questions in language acquisition, and the ability of machines to achieve human-like linguistic competence.


\section{Conclusions}
We present a comparative analysis of the usage of "red" and "green" color terms in linguistic productions of red-green CB individuals and their NS counterparts. We show that color-blind speakers use these terms in less predictable contexts, and in linguistic environments evoking mental image to a lower extent. We believe that this study, along with the released dataset, helps better understanding of the effect of sensory experience on our language, and facilitates future research in this field.

\section{Ethical Considerations}
\label{sec:ethical}
We use publicly available data to study how conceptual perception of colors by color-blind individuals is reflected in their spontaneous linguistic productions. The use of publicly available data from social media platforms, such as Reddit, may raise normative and ethical concerns. These concerns are extensively studied by the research community as reported in e.g., \citet{proferes2021studying}. Here we address two main concerns. (1) Anonymity: 
Data used for this research can only be associated with participants' user IDs, which, in turn, cannot be linked to any identifiable information. Additionally, this study uses the self-reported color blindness attribute, and does not infer any personal or demographic trait.
(2) Consent: \citet{jagfeld2021understanding} debated the need to obtain informed consent for using social media data mainly because it is not straightforward to determine if posts pertain to a public or private context. Ethical guidelines for social media research \citep{benton2017ethical} and practice in comparable research projects \citep{ahmed2017using}, as well as \href{https://www.redditinc.com/policies/user-agreement-september-12-2021}{Reddit's terms of use}, regard it as acceptable to waive explicit consent if users’ anonymity is protected.

\section{Limitations}
\label{sec:limitations}
The main limitation of our work stems from the difficulty to tease apart literal vs. figurative usages of color terms in the collected data. Certain expressions are inevitably ambiguous since they may be interpreted both literally and idiomatically; e.g., "green light" can refer metaphorically to a permission to go ahead, but also can literally mean a traffic light. However, while some of our filtered fixed expressions have a \textit{possible} literal reading, typically many fewer have a \textit{common} literal reading: these findings are consistent with those of earlier work on idiomatic expressions; for example, \cite{fazly2009unsupervised} found that for 2/3 of the potentially-idiomatic expressions in their token dataset -- i.e., phrases that could be used with either an idiomatic or literal meaning -- over 75\% of their usages were in an idiomatic reading. 

While perfect distinction of fixed usages in impractical, we believe that our approach (Section~\ref{sec:fixed-exp}) largely addresses this point by excluding usages that have a \textit{common} fixed interpretation. Notably, when skipping this filtering step (i.e., considering all phrases with color terms), the results exhibit similar comparative patterns.

\section*{Acknowledgements}
We are thankful to the anonymous reviewers and the meta reviewer for their constructive feedback. We are indebtedly grateful to Suzanne Stevenson for invaluable discussions and useful comments. In addition, we would like to thank Samuel Ackerman, Hanan Singer and Johanna Panigutti for their kind help with earlier versions of this work.

\bibliography{main}

\begin{thebibliography}{43}
\expandafter\ifx\csname natexlab\endcsname\relax\def\natexlab#1{#1}\fi

\bibitem[{Ahmed et~al.(2017)Ahmed, Bath, and Demartini}]{ahmed2017using}
Wasim Ahmed, Peter~A Bath, and Gianluca Demartini. 2017.
\newblock \href
  {https://www.emerald.com/insight/content/doi/10.1108/S2398-601820180000002004/full/html}
  {Using twitter as a data source: An overview of ethical, legal, and
  methodological challenges}.
\newblock \emph{The ethics of online research}, 2:79--107.

\bibitem[{Andersen et~al.(1984)Andersen, Dunlea, and
  Kekelis}]{andersen1984blind}
Elaine~S Andersen, Anne Dunlea, and Linda~S Kekelis. 1984.
\newblock \href
  {https://www.cambridge.org/core/journals/journal-of-child-language/article/abs/blind-childrens-language-resolving-some-differences/97A221A92453926A733277CD52DB63E4}
  {Blind children's language: Resolving some differences}.
\newblock \emph{Journal of Child language}, 11(3):645--664.

\bibitem[{Argamon et~al.(2003)Argamon, Koppel, Fine, and
  Shimoni}]{argamon2003gender}
Shlomo Argamon, Moshe Koppel, Jonathan Fine, and Anat~Rachel Shimoni. 2003.
\newblock \href
  {https://www.degruyter.com/document/doi/10.1515/text.2003.014/html} {Gender,
  genre, and writing style in formal written texts}.
\newblock \emph{Text-The Hague Then Amsterdam Then Berlin-}, 23(3):321--346.

\bibitem[{Arthur et~al.(2007)Arthur, Johnson, and Young}]{arthur2007gender}
Heather Arthur, Gail Johnson, and Adena Young. 2007.
\newblock \href
  {https://www.ingentaconnect.com/content/sbp/sbp/2007/00000035/00000006/art00009}
  {Gender differences and color: Content and emotion of written descriptions}.
\newblock \emph{Social Behavior and Personality: an international journal},
  35(6):827--834.

\bibitem[{Athanasopoulos(2011)}]{athanasopoulos2011color}
Panos Athanasopoulos. 2011.
\newblock \href
  {https://www.taylorfrancis.com/chapters/edit/10.4324/9780203836859-18/color-bilingual-cognition-panos-athanasopoulos}
  {Color and bilingual cognition}.
\newblock In \emph{Language and bilingual cognition}, pages 255--276.
  Psychology Press.

\bibitem[{Barsalou et~al.(2008)}]{barsalou2008grounded}
Lawrence~W Barsalou et~al. 2008.
\newblock \href
  {http://barsaloulab.org/Online_Articles/2008-Barsalou-ARP-grounded_cognition.pdf}
  {Grounded cognition}.
\newblock \emph{Annual review of psychology}, 59(1):617--645.

\bibitem[{Bengio et~al.(2021)Bengio, Lecun, and Hinton}]{bengio2021deep}
Yoshua Bengio, Yann Lecun, and Geoffrey Hinton. 2021.
\newblock \href {https://dl.acm.org/doi/abs/10.1145/3448250} {Deep learning for
  ai}.
\newblock \emph{Communications of the ACM}, 64(7):58--65.

\bibitem[{Benton et~al.(2017)Benton, Coppersmith, and
  Dredze}]{benton2017ethical}
Adrian Benton, Glen Coppersmith, and Mark Dredze. 2017.
\newblock \href {https://aclanthology.org/W17-1612/} {Ethical research
  protocols for social media health research}.
\newblock In \emph{Proceedings of the first ACL workshop on ethics in natural
  language processing}, pages 94--102.

\bibitem[{Bojanowski et~al.(2017)Bojanowski, Grave, Joulin, and
  Mikolov}]{bojanowski2016enriching}
Piotr Bojanowski, Edouard Grave, Armand Joulin, and Tomas Mikolov. 2017.
\newblock \href {https://doi.org/10.1162/tacl_a_00051} {Enriching word vectors
  with subword information}.
\newblock \emph{Transactions of the Association for Computational Linguistics},
  5:135--146.

\bibitem[{Chuang et~al.(2008)Chuang, Stone, and
  Hanrahan}]{chuang2008probabilistic}
Jason Chuang, Maureen Stone, and Pat Hanrahan. 2008.
\newblock \href {http://idl.cs.washington.edu/files/2008-ColorNames-CIC.pdf} {A
  probabilistic model of the categorical association between colors}.
\newblock In \emph{Color and Imaging Conference}, volume 2008, pages 6--11.
  Society for Imaging Science and Technology.

\bibitem[{Cortese and Fugett(2004)}]{cortese2004imageability}
Michael~J Cortese and April Fugett. 2004.
\newblock \href {https://link.springer.com/article/10.3758/BF03195585}
  {Imageability ratings for 3,000 monosyllabic words}.
\newblock \emph{Behavior Research Methods, Instruments, \& Computers},
  36(3):384--387.

\bibitem[{Devlin et~al.(2018)Devlin, Chang, Lee, and
  Toutanova}]{devlin2018bert}
Jacob Devlin, Ming-Wei Chang, Kenton Lee, and Kristina Toutanova. 2018.
\newblock \href {https://arxiv.org/abs/1810.04805} {Bert: Pre-training of deep
  bidirectional transformers for language understanding}.
\newblock \emph{arXiv preprint arXiv:1810.04805}.

\bibitem[{Eckert and McConnell-Ginet(2013)}]{eckert2013language}
Penelope Eckert and Sally McConnell-Ginet. 2013.
\newblock \href
  {https://books.google.co.il/books?hl=en&lr=&id=jBzJhxxGEaAC&oi=fnd&pg=PR7&dq=Language+and+gender#v=onepage&q=Language%20and%20gender&f=false}
  {\emph{Language and gender}}.
\newblock Cambridge University Press.

\bibitem[{Fazly et~al.(2009)Fazly, Cook, and Stevenson}]{fazly2009unsupervised}
Afsaneh Fazly, Paul Cook, and Suzanne Stevenson. 2009.
\newblock \href {https://direct.mit.edu/coli/article-abstract/35/1/61/2003}
  {Unsupervised type and token identification of idiomatic expressions}.
\newblock \emph{Computational Linguistics}, 35(1):61--103.

\bibitem[{Foglia and Wilson(2013)}]{foglia2013embodied}
Lucia Foglia and Robert~A Wilson. 2013.
\newblock \href
  {https://wires.onlinelibrary.wiley.com/doi/abs/10.1002/wcs.1226} {Embodied
  cognition}.
\newblock \emph{Wiley Interdisciplinary Reviews: Cognitive Science},
  4(3):319--325.

\bibitem[{Francis et~al.(2021)Francis, Rabinovich, Samir, Mortensen, and
  Stevenson}]{francis2021quantifying}
David Francis, Ella Rabinovich, Farhan Samir, David Mortensen, and Suzanne
  Stevenson. 2021.
\newblock \href
  {https://direct.mit.edu/tacl/article/doi/10.1162/tacl_a_00441/108934/Quantifying-Cognitive-Factors-in-Lexical-Decline}
  {Quantifying cognitive factors in lexical decline}.
\newblock \emph{Transactions of the Association for Computational Linguistics},
  9:1529--1545.

\bibitem[{Heer and Stone(2012)}]{heer2012color}
Jeffrey Heer and Maureen Stone. 2012.
\newblock \href
  {https://dl.acm.org/doi/abs/10.1145/2207676.2208547?casa_token=m9BtBGe6IngAAAAA:6TI1pY2mSU_VzTOzVu9VB9nS7TGCQRqMY0cZ0FvkXOdkCxExl0mm94psqlOqBPrih7kOBJSUb9c1-w}
  {Color naming models for color selection, image editing and palette design}.
\newblock In \emph{Proceedings of the SIGCHI Conference on Human Factors in
  Computing Systems}, pages 1007--1016.

\bibitem[{Hollis and Westbury(2016)}]{hollis2016principals}
Geoff Hollis and Chris Westbury. 2016.
\newblock \href {https://link.springer.com/article/10.3758/s13423-016-1053-2}
  {The principals of meaning: Extracting semantic dimensions from co-occurrence
  models of semantics}.
\newblock \emph{Psychonomic bulletin \& review}, 23(6):1744--1756.

\bibitem[{Holmes(1984)}]{holmes1984women}
Janet Holmes. 1984.
\newblock \href
  {https://www.proquest.com/openview/544b5a690c9b3f7daf59df972061fbac/1?pq-origsite=gscholar&cbl=1816443}
  {'women's language': a functional approach}.
\newblock \emph{General Linguistics}, 24(3):149.

\bibitem[{Jagfeld et~al.(2021)Jagfeld, Lobban, Rayson, and
  Jones}]{jagfeld2021understanding}
Glorianna Jagfeld, Fiona Lobban, Paul Rayson, and Steven~H Jones. 2021.
\newblock \href {https://arxiv.org/pdf/2104.11612.pdf} {Understanding who uses
  reddit: Profiling individuals with a self-reported bipolar disorder
  diagnosis}.
\newblock \emph{arXiv preprint arXiv:2104.11612}.

\bibitem[{Josserand et~al.(2021)Josserand, Meeussen, Majid, and
  Dediu}]{josserand2021environment}
Mathilde Josserand, Emma Meeussen, Asifa Majid, and Dan Dediu. 2021.
\newblock \href {https://www.nature.com/articles/s41598-021-98550-3}
  {Environment and culture shape both the colour lexicon and the genetics of
  colour perception}.
\newblock \emph{Scientific reports}, 11(1):1--11.

\bibitem[{Koppel et~al.(2002)Koppel, Argamon, and
  Shimoni}]{koppel2002automatically}
Moshe Koppel, Shlomo Argamon, and Anat~Rachel Shimoni. 2002.
\newblock \href
  {https://academic.oup.com/dsh/article-abstract/17/4/401/1019830}
  {Automatically categorizing written texts by author gender}.
\newblock \emph{Literary and linguistic computing}, 17(4):401--412.

\bibitem[{Labov(1990)}]{labov1990intersection}
William Labov. 1990.
\newblock \href
  {https://www.cambridge.org/core/journals/language-variation-and-change/article/intersection-of-sex-and-social-class-in-the-course-of-linguistic-change/AAA8227B739187F5D2CBDA51EA212FD8}
  {The intersection of sex and social class in the course of linguistic
  change}.
\newblock \emph{Language variation and change}, 2(2):205--254.

\bibitem[{Lakoff(1973)}]{lakoff1973language}
Robin Lakoff. 1973.
\newblock \href
  {https://www.cambridge.org/core/journals/language-in-society/article/language-and-womans-place/F66DB3D1BB878CDD68B9A79A25B67DE6}
  {Language and woman's place}.
\newblock \emph{Language in society}, 2(1):45--79.

\bibitem[{Landau and Gleitman(2009)}]{landau2009language}
Barbara Landau and Lila~R Gleitman. 2009.
\newblock \href {https://psycnet.apa.org/record/1985-97756-000} {\emph{Language
  and experience: Evidence from the blind child}}, volume~8.
\newblock Harvard University Press.

\bibitem[{LeCun et~al.(2015)LeCun, Bengio, and Hinton}]{lecun2015deep}
Yann LeCun, Yoshua Bengio, and Geoffrey Hinton. 2015.
\newblock \href {https://www.nature.com/articles/nature14539} {Deep learning}.
\newblock \emph{nature}, 521(7553):436--444.

\bibitem[{Lewis and Lupyan(2020)}]{lewis2020gender}
Molly Lewis and Gary Lupyan. 2020.
\newblock \href {https://www.nature.com/articles/s41562-020-0918-6} {Gender
  stereotypes are reflected in the distributional structure of 25 languages}.
\newblock \emph{Nature human behaviour}, 4(10):1021--1028.

\bibitem[{Lynott et~al.(2020)Lynott, Connell, Brysbaert, Brand, and
  Carney}]{lynott2020lancaster}
Dermot Lynott, Louise Connell, Marc Brysbaert, James Brand, and James Carney.
  2020.
\newblock \href {https://link.springer.com/article/10.3758/s13428-019-01316-z}
  {The lancaster sensorimotor norms: multidimensional measures of perceptual
  and action strength for 40,000 english words}.
\newblock \emph{Behavior Research Methods}, 52(3):1271--1291.

\bibitem[{Matusevych et~al.(2018)Matusevych, Beekhuizen, and
  Stevenson}]{matusevych2018crosslinguistic}
Yevgen Matusevych, Barend Beekhuizen, and Suzanne Stevenson. 2018.
\newblock \href
  {http://www.cs.utoronto.ca/~barend/data/matusevych_et_al_2018.pdf}
  {Crosslinguistic transfer as category adjustment: Modeling conceptual color
  shift in bilingualism.}
\newblock In \emph{CogSci}.

\bibitem[{McMahan and Stone(2015)}]{mcmahan2015bayesian}
Brian McMahan and Matthew Stone. 2015.
\newblock \href
  {https://direct.mit.edu/tacl/article/doi/10.1162/tacl_a_00126/43263/A-Bayesian-Model-of-Grounded-Color-Semantics}
  {A bayesian model of grounded color semantics}.
\newblock \emph{Transactions of the Association for Computational Linguistics},
  3:103--115.

\bibitem[{Mikolov et~al.(2013)Mikolov, Chen, Corrado, and
  Dean}]{mikolov2013efficient}
Tomas Mikolov, Kai Chen, Greg Corrado, and Jeffrey Dean. 2013.
\newblock \href {https://arxiv.org/abs/1301.3781} {Efficient estimation of word
  representations in vector space}.
\newblock \emph{arXiv preprint arXiv:1301.3781}.

\bibitem[{Pennington et~al.(2014)Pennington, Socher, and
  Manning}]{pennington2014glove}
Jeffrey Pennington, Richard Socher, and Christopher~D Manning. 2014.
\newblock \href {https://aclanthology.org/D14-1162.pdf} {Glove: Global vectors
  for word representation}.
\newblock In \emph{Proceedings of the 2014 conference on empirical methods in
  natural language processing (EMNLP)}, pages 1532--1543.

\bibitem[{P{\'e}rez-Pereira and Conti-Ramsden(2013)}]{perez2013language}
Miguel P{\'e}rez-Pereira and Gina Conti-Ramsden. 2013.
\newblock \href
  {https://www.taylorfrancis.com/books/mono/10.4324/9780203776087/language-development-social-interaction-blind-children-miguel-perez-pereira-gina-conti-ramsden}
  {\emph{Language development and social interaction in blind children}}.
\newblock Psychology Press.

\bibitem[{Proferes et~al.(2021)Proferes, Jones, Gilbert, Fiesler, and
  Zimmer}]{proferes2021studying}
Nicholas Proferes, Naiyan Jones, Sarah Gilbert, Casey Fiesler, and Michael
  Zimmer. 2021.
\newblock \href
  {https://journals.sagepub.com/doi/full/10.1177/20563051211019004} {Studying
  reddit: A systematic overview of disciplines, approaches, methods, and
  ethics}.
\newblock \emph{Social Media+ Society}, 7(2):20563051211019004.

\bibitem[{Rabinovich et~al.(2020)Rabinovich, Gonen, and
  Stevenson}]{rabinovich2020pick}
Ella Rabinovich, Hila Gonen, and Suzanne Stevenson. 2020.
\newblock \href {https://arxiv.org/abs/2011.00335} {Pick a fight or bite your
  tongue: Investigation of gender differences in idiomatic language usage}.
\newblock In \emph{Proceedings of the 28th International Conference on
  Computational Linguistics}, pages 5181--5192.

\bibitem[{Rabinovich et~al.(2017)Rabinovich, Patel, Mirkin, Specia, and
  Wintner}]{rabinovich2017personalized}
Ella Rabinovich, Raj~Nath Patel, Shachar Mirkin, Lucia Specia, and Shuly
  Wintner. 2017.
\newblock \href {https://arxiv.org/abs/1610.05461} {Personalized machine
  translation: Preserving original author traits}.
\newblock In \emph{Proceedings of the 15th Conference of the European Chapter
  of the Association for Computational Linguistics}, pages 1074--1084.

\bibitem[{Sarantakis(2014)}]{sarantakis2014influence}
Nicholas~P Sarantakis. 2014.
\newblock \href
  {https://pure.northampton.ac.uk/en/publications/the-influence-of-our-native-language-on-cognitive-representations}
  {The influence of our native language on cognitive representations of colour,
  spatial relations and time}.
\newblock \emph{Journal of European Psychology Students}, 5(3):74--77.

\bibitem[{Scott et~al.(2019)Scott, Keitel, Becirspahic, Yao, and
  Sereno}]{scott2019glasgow}
Graham~G Scott, Anne Keitel, Marc Becirspahic, Bo~Yao, and Sara~C Sereno. 2019.
\newblock \href {https://link.springer.com/article/10.3758/s13428-018-1099-3}
  {The glasgow norms: Ratings of 5,500 words on nine scales}.
\newblock \emph{Behavior research methods}, 51(3):1258--1270.

\bibitem[{Shepard and Cooper(1992)}]{shepard1992representation}
Roger~N Shepard and Lynn~A Cooper. 1992.
\newblock \href
  {https://journals.sagepub.com/doi/abs/10.1111/j.1467-9280.1992.tb00006.x}
  {Representation of colors in the blind, color-blind, and normally sighted}.
\newblock \emph{Psychological science}, 3(2):97--104.

\bibitem[{Tenney et~al.(2019)Tenney, Xia, Chen, Wang, Poliak, McCoy, Kim,
  Van~Durme, Bowman, Das et~al.}]{tenney2019you}
Ian Tenney, Patrick Xia, Berlin Chen, Alex Wang, Adam Poliak, R~Thomas McCoy,
  Najoung Kim, Benjamin Van~Durme, Samuel~R Bowman, Dipanjan Das, et~al. 2019.
\newblock \href {https://arxiv.org/abs/1905.06316} {What do you learn from
  context? probing for sentence structure in contextualized word
  representations}.
\newblock \emph{arXiv preprint arXiv:1905.06316}.

\bibitem[{Tsvetkov et~al.(2013)Tsvetkov, Mukomel, and
  Gershman}]{tsvetkov2013cross}
Yulia Tsvetkov, Elena Mukomel, and Anatole Gershman. 2013.
\newblock \href {https://aclanthology.org/W13-0906.pdf} {Cross-lingual metaphor
  detection using common semantic features}.
\newblock In \emph{Proceedings of the First Workshop on Metaphor in NLP}, pages
  45--51.

\bibitem[{Vaswani et~al.(2017)Vaswani, Shazeer, Parmar, Uszkoreit, Jones,
  Gomez, Kaiser, and Polosukhin}]{vaswani2017attention}
Ashish Vaswani, Noam Shazeer, Niki Parmar, Jakob Uszkoreit, Llion Jones,
  Aidan~N Gomez, {\L}ukasz Kaiser, and Illia Polosukhin. 2017.
\newblock \href
  {https://proceedings.neurips.cc/paper/2017/hash/3f5ee243547dee91fbd053c1c4a845aa-Abstract.html}
  {Attention is all you need}.
\newblock \emph{Advances in neural information processing systems}, 30.

\bibitem[{Wong(2011)}]{wong2011color}
Bang Wong. 2011.
\newblock \href {https://www.nature.com/articles/nmeth.1618} {Points of view:
  Color blindness}.
\newblock \emph{Nature methods}, 8(6):441.

\end{thebibliography}
\bibliographystyle{acl_natbib}

\appendix

\section{Appendix}
\label{sec:appendix}

\subsection{List of Excluded Subreddits}
This list of topical color-related threads (subreddits) could have potentially introduced a bias to our study, and therefore were excluded from data collection and analysis. We used an exhaustive list of subreddits including the term `color' in their name:

\noindent
\texttt{r/color \\
r/ColorBlind \\
r/ColorBlindGamers \\
r/colorblindmemes \\
r/colorblindness \\
r/ColorBlindnessIsFun \\
r/ColorGrading \\
r/colorists \\
r/Colorization \\ 
r/colorizationrequests \\
r/colorizebot \\
r/colorize\_bw\_photos \\
r/ColorizedHistory \\
r/colorpie \\
r/colorsbot \\
r/ColorshopBattles \\
r/Colorslash \\
r/ColorThisSpace \\
r/Colourblind \\
r/ImaginaryColorscapes \\
r/thecolorless \\
r/UnitedColors \\
r/Shitty\_Watercolour \\
r/WatercolorChallenge \\
r/Whatcoloristhis \\
r/Watercolor}

\subsection{Fixed Expressions with Color Terms Excluded from this Study}
Table~\ref{tbl:fixed-expressions} reports the list of fixed constructions with color terms that were excluded from this work. We refer to some limitations related to filtering fixed expressions with color terms in Section~\ref{sec:limitations}.

\begin{table}[h!]
\centering
\resizebox{\columnwidth}{!}{
\begin{tabular}{l|l}
term & nouns used in fixed constructions \\ \hline
red & \specialcell{alert, bull, cross, dot, flag, flags, herring, line, \\ meat, pill, redemption, sox, tape, wedding, \\ wings, zone} \\ \hline
green & \specialcell{bay, card, cards, day, deal, earth, energy, \\ lantern, light, line, mile, party, screen, text} \\ \hline
black & \specialcell{box, flag, friday, guy, guys, hole, holes, magic, \\ man, market, men, metal, mirror, ops, panther, \\ people, person, widow, woman, women} \\ \hline
white & \specialcell{dude, guy, guys, house, knight, male, males, \\ man, men, nationalists, noise, people, person, \\ privilege, sox, supremacist, supremacists, \\ supremacy, walkers, women} \\ \hline
blue & \specialcell{balls, blood, cheese, collar, jackets, jays, light, \\ line, moon, shell, state, states, team, whale} \\ \hline
brown & \specialcell{guy, people, recluse, switches} \\ \hline
gr[ae]y & \specialcell{area, areas, chapter, cup, goo, goose, jedi, \\ knights, man, market, matter, poupon, video, \\ wardens, wind, worm, zone} \\ \hline
yellow & \specialcell{fever, jacket, jackets, journalism, \\ pages, submarine} \\ \hline
pink & \specialcell{album, eye, floyd, guy, mast, panther, pistols, \\ ranger, slime, slip, song, tax} \\ \hline

purple & \specialcell{drank, gang, haze, heart, hearts, line, link, \\ links, man, rain} \\

\end{tabular}
}
\caption{The list of fixed expressions with color terms excluded from this study.}
\label{tbl:fixed-expressions}
\end{table}

\subsection{Experimental Results for Control Set \#2: NS Population Split}

Tables~\ref{tbl:frequencies-ns}, \ref{tbl:all-img-ns} and \ref{tbl:cwp-ns} report the results. Significant differences in frequencies are shown for "red" and "social" between the two sets (Table~\ref{tbl:frequencies-ns}), "black" is the only color term exhibiting signinficant difference between the two groups with very low effect size (Cohen's-$d\text{=}0.038$) -- this difference can be attributed to the very large sample size. BERT ranks predictions differ for the "loose" and "royal" adjectives; again, with very low effect sizes.

\begin{table}[h!]
\centering
\resizebox{\columnwidth}{!}{
\begin{tabular}{l|rl|rr}
color term & NS1 freq  & NS2 freq & NS1 ratio  & NS2 ratio \\ \hline
red     & 2.07e-4 & 8.08e-5*    & 0.170 & 0.165 \\
green   & 1.16e-5 & 1.49e-5     & 0.087 & 0.085 \\ \hline
black   & 3.40e-5 & 4.12e-5     & 0.256 & 0.254 \\
white   & 2.74e-5 & 3.68e-5     & 0.207 & 0.209 \\
blue    & 1.57e-5 & 1.45e-5     & 0.118 & 0.119 \\
brown   & 6.67e-5 & 6.02e-5     & 0.050 & 0.050 \\
gr[ae]y & 2.75e-5 & 2.30e-5     & 0.032 & 0.033 \\
yellow  & 4.63e-5 & 4.89e-5     & 0.034 & 0.035 \\
pink    & 2.91e-6 & 3.67e-6     & 0.022 & 0.023 \\
purple  & 3.24e-6 & 3.01e-6     & 0.024 & 0.023 \\ \hline
\textbf{total} &  &  & 1.0 & 1.0 \\ \hline \hline
hot     & 2.71e-5 & 1.57e-5     & 0.210 & 0.208 \\
social  & 3.75e-5 & 2.06e-5*    & 0.291 & 0.287 \\
clear   & 2.81e-5 & 2.12e-5     & 0.218 & 0.219 \\
tiny    & 1.08e-5 & 2.09e-5     & 0.083 & 0.086 \\
flat    & 9.74e-6 & 7.47e-6     & 0.075 & 0.078 \\
loose   & 5.28e-6 & 4.19e-6     & 0.041 & 0.039 \\
petty   & 2.58e-6 & 1.39e-6     & 0.020 & 0.021 \\
clever  & 2.63e-6 & 3.77e-6     & 0.020 & 0.022 \\
royal   & 2.72e-6 & 1.78e-6     & 0.021 & 0.019 \\
lame    & 2.28e-6 & 2.07e-6     & 0.017 & 0.017 \\ \hline
\textbf{total} &  &  & 1.0 & 1.0
\end{tabular}
}
\caption{Relative frequencies (left) and relative ratios (right) of color terms in the language of CB and NS population. Statistical significance of the differences was tested using a two-proportion z-test; "*" indicates significant difference at the level of $p\text{<}.01$.}
\label{tbl:frequencies-ns}
\end{table}

\begin{table}[h!]
\centering
\resizebox{\columnwidth}{!}{
\begin{tabular}{l|rr|rr|r}
& \multicolumn{2}{c|}{NS1} & \multicolumn{2}{c|}{NS2} & \\ \hline
term & \# sent      & M(img) & \# sent  & M(img) & Cohen's d \\ \hline
red     & 10,358    & 0.774 & 47,512    & 0.772\hspace{0.20cm}  & -0.009 \\
green   & 4,795     & 0.730 & 21,364    & 0.738\hspace{0.20cm}  & 0.036 \\ \hline
black   & 17,288    & 0.722 & 76,879    & 0.730*                & 0.038 \\
white   & 13,618    & 0.719 & 61,705    & 0.718\hspace{0.20cm}  & -0.002 \\
blue    & 7,045     & 0.764 & 31,588    & 0.765\hspace{0.20cm}  & 0.004 \\
brown   & 2,510     & 0.797 & 11,038    & 0.801\hspace{0.20cm}  & 0.016 \\
gr[ae]y & 1,542     & 0.731 & 7,422     & 0.739\hspace{0.20cm}  & 0.042 \\
yellow  & 2,038     & 0.774 & 9,054     & 0.772\hspace{0.20cm}  & -0.007 \\
pink    & 1,300     & 0.779 & 5,946     & 0.788\hspace{0.20cm}  & 0.052 \\
purple  & 1,385     & 0.727 & 5,946     & 0.746\hspace{0.20cm}  & 0.088 \\ \hline \hline
hot     & 12,406    & 0.760 & 55,056    & 0.760\hspace{0.20cm}  & 0.003 \\
social  & 23,152    & 0.448 & 102,419   & 0.445\hspace{0.20cm}  & -0.016 \\
clear   & 6,339     & 0.535 & 28,157    & 0.531\hspace{0.20cm}  & -0.018 \\
tiny    & 6,559     & 0.609 & 30,082    & 0.611\hspace{0.20cm}  & 0.007 \\
flat    & 3,594     & 0.677 & 16,735    & 0.677\hspace{0.20cm}  & 0.002 \\
loose   & 1,693     & 0.635 & 7,441     & 0.652\hspace{0.20cm}  & 0.069 \\
petty   & 1,109     & 0.504 & 5,227     & 0.505\hspace{0.20cm}  & 0.006 \\
clever  & 1,079     & 0.478 & 5,130     & 0.486\hspace{0.20cm}  & 0.034 \\
royal   & 1,402     & 0.710 & 5,733     & 0.694*                & -0.093 \\
lame    & 938       & 0.538 & 4,399     & 0.538\hspace{0.20cm}  & 0.001 \\
\end{tabular}
}
\caption{Mean imageability scores of nouns modified by color terms and the control set \#1 adjectives in the control \#2 NS split. "*" indicates significant difference at the level of $p\text{<}.01$.}
\label{tbl:all-img-ns}
\end{table}

\begin{table}[h!]
\centering
\resizebox{\columnwidth}{!}{
\begin{tabular}{l|rr|rr|r}
& \multicolumn{2}{c|}{NS1} & \multicolumn{2}{c|}{NS2} & \\ \hline
term & \# sent      & M(rank) & \# sent  & M(rank) & Cohen's d \\ \hline
red     & 15,472    & 71.7  & 70,177    & 75.3\hspace{0.20cm}   & -0.022 \\
green   & 8,500     & 103.5 & 37,775    & 107.8\hspace{0.20cm}  & -0.022 \\ \hline
black   & 24,492    & 54.6  & 110,205   & 54.0\hspace{0.20cm}   & 0.004 \\
white   & 21,048    & 67.2  & 96,331    & 69.0\hspace{0.20cm}   & -0.012 \\
blue    & 11,508    & 100.9 & 52,367    & 97.8\hspace{0.20cm}   & 0.016 \\
brown   & 4,870     & 111.6 & 22,440    & 109.9\hspace{0.20cm}  & 0.008 \\
grey    & 2,876     & 158.5 & 13,756    & 151.1\hspace{0.20cm}  & 0.030 \\
yellow  & 3,207     & 102.2 & 14,716    & 103.1\hspace{0.20cm}  & -0.005 \\
pink    & 1,989     & 106.0 & 9,202     & 112.7\hspace{0.20cm}  & -0.035 \\
purple  & 2,001     & 154.7 & 8,812     & 144.9\hspace{0.20cm}  & 0.046 \\ \hline \hline
hot     & 20,637    & 55.2  & 91,331    & 56.4\hspace{0.20cm}   & -0.009 \\ 
social  & 24,271    & 27.7  & 107,728   & 28.0\hspace{0.20cm}   & -0.003 \\
clear   & 21,464    & 47.0  & 97,639    & 45.3\hspace{0.20cm}   & 0.014 \\
tiny    & 7,321     & 70.4  & 33,591    & 72.9\hspace{0.20cm}   & -0.017 \\
flat    & 6,950     & 83.7  & 32,030    & 90.3\hspace{0.20cm}   & -0.038 \\
loose   & 3,730     & 143.3 & 16,098    & 129.7*                & 0.061 \\
petty   & 1,699     & 152.6 & 8,224     & 155.3\hspace{0.20cm}  & -0.012 \\
clever  & 2,151     & 105.2 & 10,391    & 106.1\hspace{0.20cm}  & -0.005 \\
royal   & 1,699     & 77.5  & 6,910     & 85.6*                 & -0.045 \\
lame    & 1,761     & 179.9 & 8,231     & 184.4\hspace{0.20cm}  & -0.020 \\
\end{tabular}
}
\caption{Mean BERT rank predictions for a masked term. Results for both color terms and control set \#1 adjectives in control set \#2 NS split are reported. "*" indicates significant difference at the level of $p\text{<}.01$.}
\label{tbl:cwp-ns}
\end{table}

\end{document}